\documentclass[runningheads]{llncs}

\usepackage{graphicx}
\usepackage{booktabs}

\usepackage[accsupp]{axessibility}  

\usepackage{hyperref}

\usepackage{orcidlink}

\usepackage{graphicx}
\usepackage{amsmath,amssymb} 
\usepackage{color}
\usepackage{booktabs}
\usepackage{multirow}
\usepackage{booktabs}      
\usepackage{multirow}      
\usepackage{array}         
\usepackage{threeparttable}
\usepackage{arydshln}      

\usepackage{xspace}
\usepackage{xcolor}
\usepackage{wrapfig}
\usepackage{caption}

\begin{document}

\title{From Imitation to Intuition: Intrinsic Reasoning for Open-Instance Video Classification} 

\titlerunning{From Imitation to Intuition}

\authorrunning{K. Zhang et al.}

\author{
Ke Zhang\inst{1,2}\thanks{Equal contribution.} \and
Xiangchen Zhao\inst{2}\textsuperscript{\textasteriskcentered}\and
Yunjie Tian\inst{2} \and
Jiayu Zheng\inst{2} \and
Vishal M. Patel\inst{1} \and
Di Fu\inst{2}
}

\institute{
Johns Hopkins University \and
ByteDance Inc
}

\newcommand{\model}{DeepIntuit\xspace}

\maketitle

\begin{figure}[!h]
    \centering
    \includegraphics[width=\textwidth]{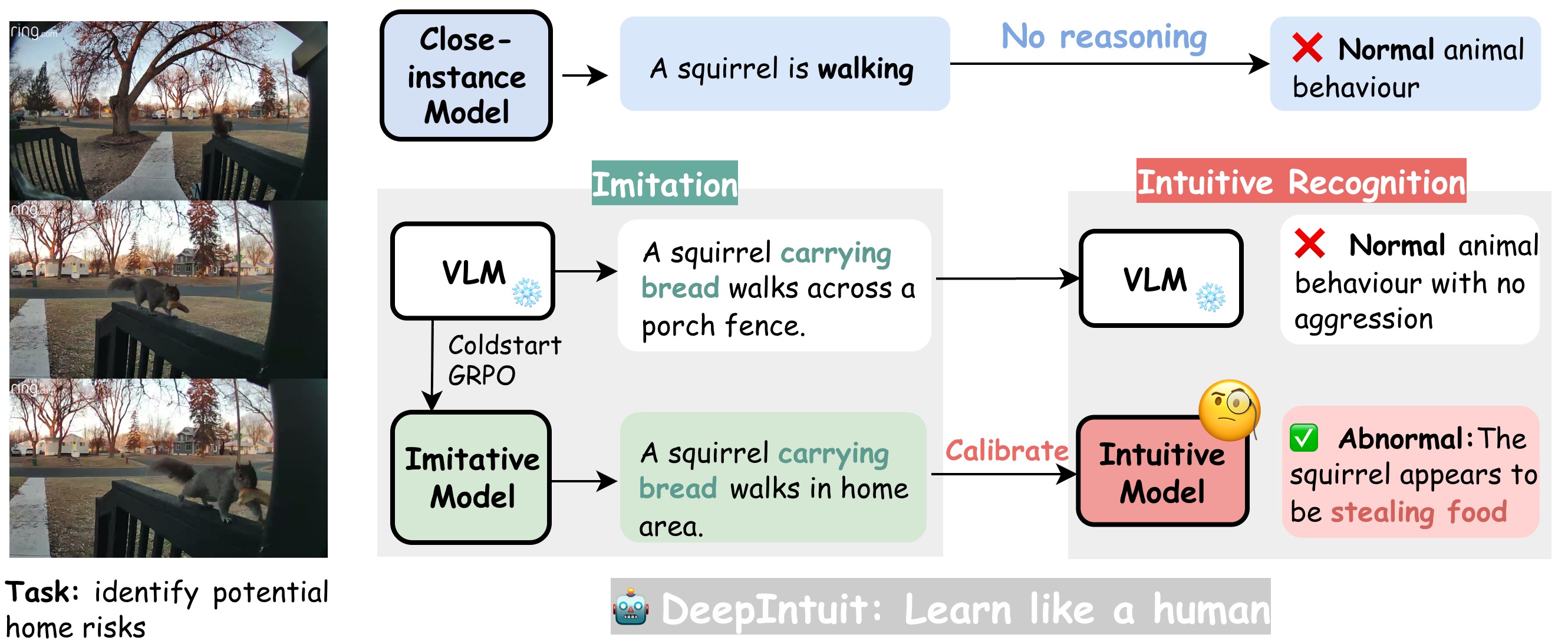}
    \caption{{\textbf{Overview of \model.} Unlike conventional classifiers that rely on direct input-to-label mapping, \model evolves open-instance video classification from imitation to intuition. Through staged training, it develops intrinsic reasoning that enables stable and calibrated decisions.}}
    \label{fig:teaser}
    \vspace{-25pt}
\end{figure}

\begin{abstract}

Conventional video classification models, acting as effective imitators, excel in scenarios with homogeneous data distributions.
However, real-world applications often present an open-instance challenge, where intra-class variations are vast and complex, beyond existing benchmarks.
While traditional video encoder models struggle to fit these diverse distributions, vision-language models (VLMs) offer superior generalization but have not fully leveraged their reasoning capabilities (intuition) for such tasks.
In this paper, we bridge this gap with an intrinsic reasoning framework that evolves open-instance video classification from imitation to intuition.
Our approach, namely \model, begins with a cold-start supervised alignment to initialize reasoning capability, followed by refinement using Group Relative Policy Optimization (GRPO) to enhance reasoning coherence through reinforcement learning.
Crucially, to translate this reasoning into accurate classification, \model then introduces an intuitive calibration stage. 
In this stage, a classifier is trained on this intrinsic reasoning traces generated by the refined VLM, ensuring stable knowledge transfer without distribution mismatch.
Extensive experiments demonstrate that for open-instance video classification, \model benefits significantly from transcending simple feature imitation and evolving toward intrinsic reasoning.
Our project is available at \url{https://bwgzk-keke.github.io/DeepIntuit/}.

\keywords{Open-instance, Video classification, Intrinsic reasoning, GRPO}
\end{abstract}

\begin{figure}[t]
    \centering
    \includegraphics[width=\textwidth]{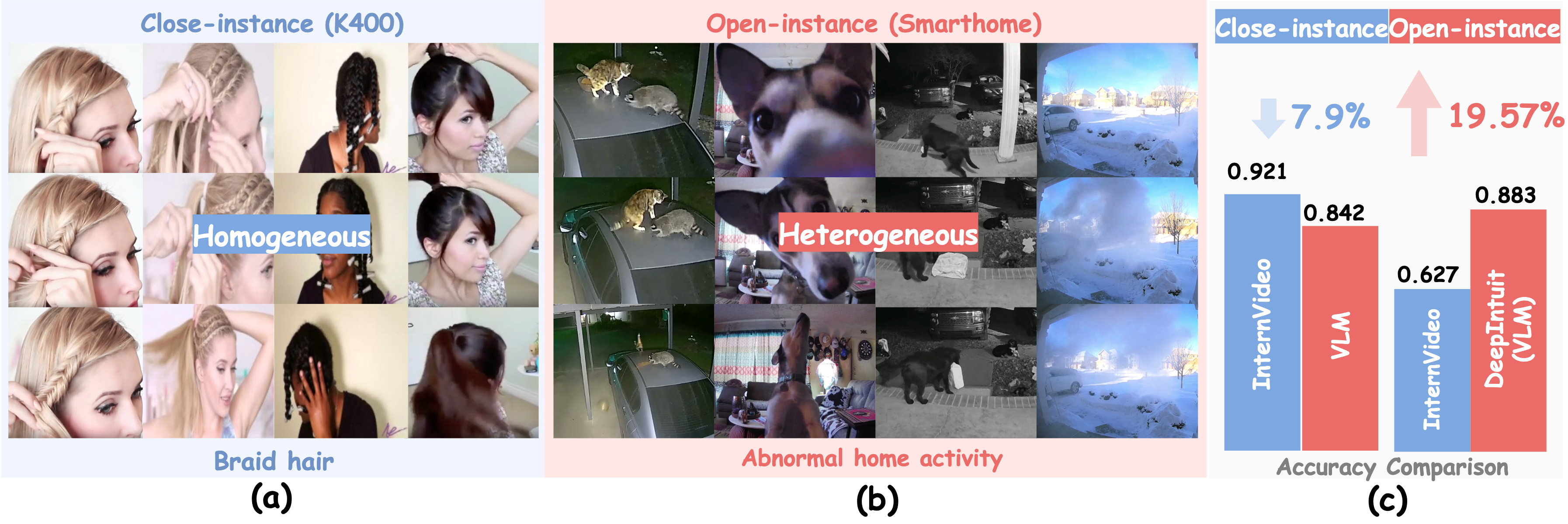}
    \caption{\textbf{Close-instance vs.\ open-instance video classification.} (a) Close-instance benchmarks have relatively homogeneous intra-class distributions. (b) Open-instance settings exhibit broader, open-ended intra-class variation that better reflects real-world data. (c) Consequently, conventional video encoders fit close-instance data well but struggle to generalize, whereas VLMs with stronger semantic priors are more robust in the open-instance regime.}
    \label{fig:motivation}
    \vspace{-10pt}
\end{figure}

\section{Introduction}
\label{sec:intro}
{
Open-instance video classification poses a fundamentally different challenge from traditional video classification, as shown in Figure~\ref{fig:motivation}. In this setting, the label space remains fixed, but each class exhibits large and open-ended intra-class variation in appearance, motion, context, and semantics. Unlike traditional benchmarks~\cite{goyal2017something,kay2017kinetics}, where train and test data often share relatively homogeneous distributions and models can succeed through simple fitting, real-world video classification must generalize across much broader instance diversity. As a result, conventional video encoders that rely on direct feature fitting often struggle (see Figure~\ref{fig:motivation}), while vision-language models (VLMs), with stronger semantic priors from large-scale multimodal pretraining, provide a more suitable foundation.
}

{
However, VLMs should not be treated as conventional classifiers. A straightforward approach is to fine-tune a VLM to output a single class token for each video~\cite{daillm,shi2024math}, reducing classification to a direct input-to-label mapping. Yet this is poorly suited to open-instance settings, where robust decisions require more than surface-level fitting. By bypassing the model’s internal semantic analysis, such optimization often leads to poor calibration and can even damage the VLM’s original open-ended understanding and question-answering ability, pushing it toward collapsed task-specific biases. The key challenge is therefore to turn the VLM’s latent reasoning capacity into reliable classification behavior without sacrificing its generative competence.
}

{
Recent advances in reinforcement learning (RL)–based reasoning~\cite{grpo_cite,zhang2025r1,dsr1_cite,openaio1_cite,team2025kimi,wei2025open} suggest a promising direction. Rather than directly enforcing an input-to-label mapping, RL-based reasoning encourages models to externalize and refine intermediate reasoning. Prior work shows that strong reasoning behavior arises from structured internal cognitive patterns rather than scale alone~\cite{yu2024rlhf,gandhi2025cognitive,zhao2025echo}. By optimizing for structured and interpretable rationales, RL promotes behaviors such as intermediate verification and hypothesis revision~\cite{zhao2025echo}. Moreover, the shift from preference-based RLHF~\cite{ouyang2022training} to rule-grounded reward optimization such as RLVR~\cite{dsr1_cite} further improves stability by reducing reward hacking and spurious reward exploitation~\cite{rewardhack,yu2025unhackable}. These advances make RL-based reasoning a natural way to elicit the latent reasoning ability of VLMs.
}

{
However, directly applying RL-trained reasoning models to open-instance video classification remains brittle. Even when RL improves the reasoning process, the resulting model can still be unreliable for final classification: it may produce plausible intermediate reasoning while its final prediction remains poorly aligned with actual correctness. The core issue is that reasoning traces are often treated as final evidence, without an explicit step to calibrate how they should support the final decision. As a result, stronger reasoning does not automatically yield better classification, and errors or overconfident judgments can be passed directly to the output layer.
}

{
Existing methods such as Chain-of-Thought (CoT) prompting and rationale supervision improve the visibility of reasoning~\cite{huang2022large,wei2022chain,wang2022self}, but they still mainly treat reasoning traces as supervision signals rather than something to be calibrated for classification. Thus, they improve interpretability, but do not fully solve the reliability problem in open-instance video classification.
}

{
In contrast, we propose an intrinsic reasoning framework that evolves open-instance video classification from imitation to intuition. Rather than reducing a VLM to a single-step classifier, \model develops its latent reasoning ability and translates it into reliable classification behavior. Specifically, \model consists of three stages: \textbf{1)} cold-start supervised alignment, which establishes an initial reasoning prior using reasoning data; \textbf{2)} GRPO-based reinforcement learning, which enhances the reasoning process; and \textbf{3)} intuitive calibration, which trains a classifier on intrinsic reasoning traces generated by the same VLM. This design decouples reasoning from final decision making, allowing classification to build on internal reasoning while avoiding the instability of directly treating reasoning outputs as final predictions.
}

{
This intrinsic reasoning framework is essential for stable open-instance video classification. The supervised alignment and RL stages strengthen reasoning capability, while the calibration stage turns that reasoning into robust and calibrated decisions. Importantly, training the calibration model on reasoning traces generated by the same model preserves distribution consistency and avoids performance degradation caused by mismatched reasoning and decision layers. Through extensive experiments on diverse real-world open-instance classification scenarios, we show that evolving toward intrinsic reasoning, rather than relying on direct classification or naive RL deployment alone, leads to substantially better robustness and generalization under complex intra-class variation.
}

Our contributions are threefold:
\begin{itemize}
\item We introduce an intrinsic reasoning framework that evolves open-instance video classification from imitation to intuition.

\item We show that reinforcement learning improves reasoning quality, but robust classification further requires an explicit intuitive calibration stage to align reasoning with final decisions.

\item We demonstrate through extensive experiments that distribution-consistent calibration, built on intrinsic reasoning traces from the same refined VLM, is essential for stable and robust open-instance video classification.
\end{itemize}

\section{Related Work}
\subsection{Cognitive Patterns in LLMS}
Recent works~\cite{yu2024rlhf,gandhi2025cognitive,zhao2025echo} indicate that advanced reasoning in large models stems from the emergence of structured internal strategies rather than scale alone. These strategies resemble human-style problem solving, such as reconsidering earlier assumptions when contradictions arise, validating intermediate conclusions, decomposing complex tasks into smaller objectives, and reasoning backward from a target goal~\cite{zhao2025echo}. Together, these mechanisms function as an implicit reasoning scaffold, enabling stable multi-step inference.
The transition from preference-based reinforcement learning (RLHF)~\cite{ouyang2022training} to rule-grounded reward optimization (RLVR)~\cite{dsr1_cite} has further strengthened these behaviors, significantly enhancing reasoning performance in large models~\cite{dsr1_cite,openaio1_cite}. When correctness is determined by objective criteria rather than learned reward estimators, models are less prone to exploiting spurious signals, reducing reward hacking risks~\cite{rewardhack,yu2025unhackable}. This shift has been shown to stabilize large-scale training and encourage the consistent activation of structured reasoning routines~\cite{yu2024rlhf,gandhi2025cognitive,zhao2025echo}.
\subsection{VLM Cognitive Behaviors}
The idea that reasoning structures can transfer across modalities has motivated recent multimodal research~\cite{liu2025x,wei2025open,hu2025open}. 
Multimodal reinforcement learning is particularly well-suited to verifiable supervision because visual tasks are naturally grounded in objective perceptual signals~\cite{chen2023shikra,yu2025perception}. Nevertheless, early multimodal reinforcement approaches predominantly relied on RLHF-style learned reward models~\cite{wang2024mdpo,zhu2025perpo,zhu2024self}. More recent work, inspired by RLVR’s success in language reasoning~\cite{dsr1_cite,team2025kimi}, incorporates rule-based objectives into multimodal training~\cite{zhang2025think}. For instance, Perception-R1~\cite{yu2025perception} introduces explicit perceptual metrics such as IoU and geometric distance to improve grounding quality. Other approaches, including R1-OneVision~\cite{yang2025r1} and VLAA-Thinking~\cite{chen2025sft}, construct enriched reasoning trajectories via multi-stage distillation and synthesis pipelines. Additionally, ReVisual-R1 demonstrates that a language-only initialization can serve as a strong starting point for subsequent visual reasoning adaptation~\cite{revisualr1}.

\subsection{Video Cognitive Reasoning}
Although Vision-Language Models (VLMs)~\cite{bai2025qwen2,comanici2025gemini,openai2025gpt4omini} generalize well on standard benchmarks, their performance often degrades under distribution shifts where high-level labels fail to capture domain-specific semantics~\cite{kay2017kinetics}. Direct supervised fine-tuning (SFT) frequently struggles to bridge this gap, leading to overfitting or weak reasoning~\cite{xu2025rb}. The complexity of open-world video distributions~\cite{zhang2025endless,xu2025freevis}, characterized by large intra-class variation~\cite{wang2024multihateclip,zhao2025smarthome}, further limits the effectiveness of simple label supervision.
Recent work addresses this challenge through self-improvement paradigms that leverage model-generated reasoning. Chain-of-Thought (CoT) prompting introduces intermediate inference steps to enhance reasoning~\cite{wei2022chain}, while subsequent studies incorporate self-consistency and reasoning-guided training to provide richer supervision beyond categorical labels~\cite{wang2022self,huang2022large}. In multimodal settings, rationale distillation from stronger teacher models transfers reasoning knowledge to smaller models~\cite{zhang2023video}, and reflective self-training further improves robustness through iterative reasoning refinement~\cite{cheng2024vision}. More recently, reasoning-centric training has been explored for video understanding by explicitly modeling structured reasoning processes~\cite{park2025deepvideo,cao2025videominer,zhang2025think}.

\begin{figure}[t]
    \centering
    \includegraphics[width=\textwidth]{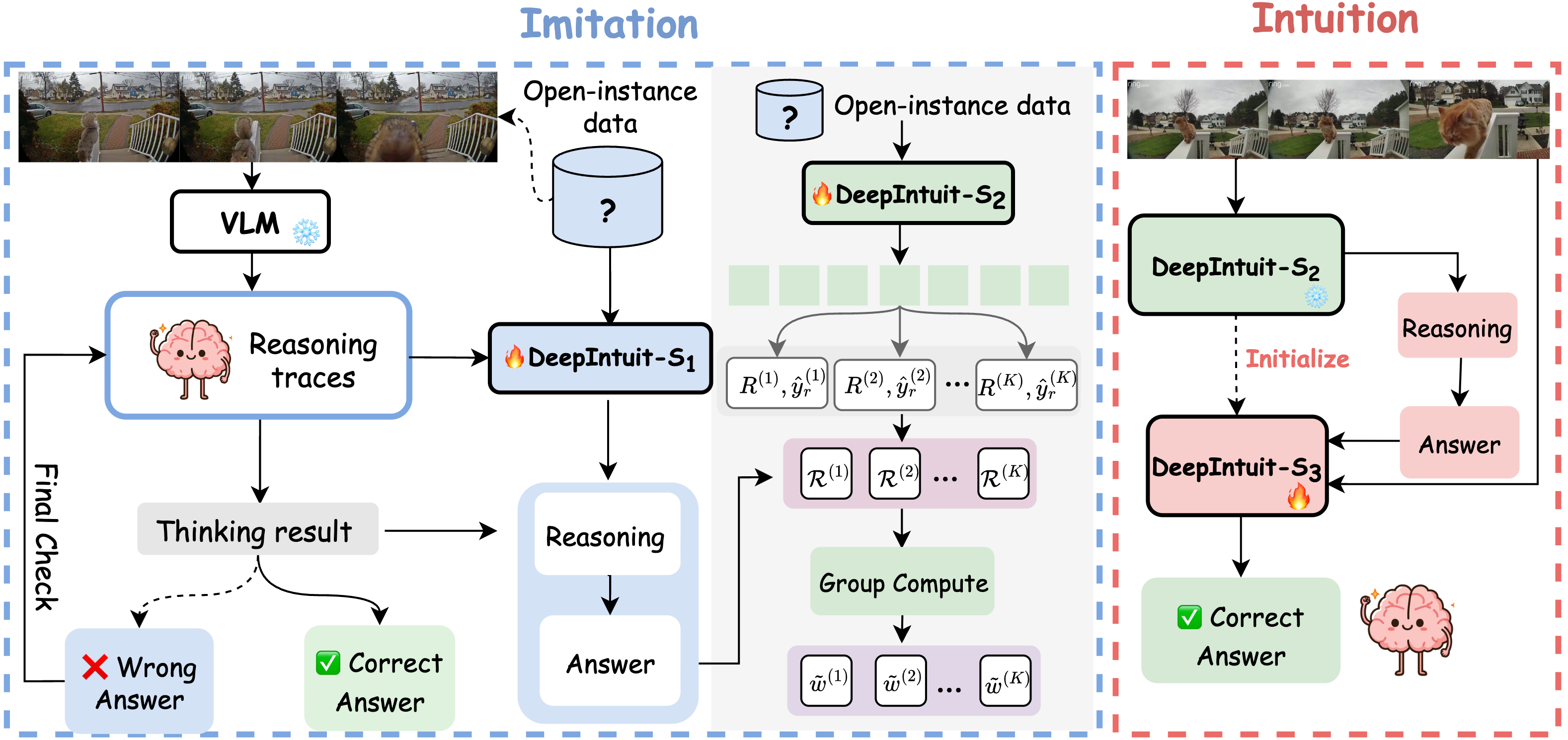}
    \caption{\textbf{Pipeline of \model.} The framework follows three stages: (1) cold-start supervised alignment for initializing reasoning capability, (2) GRPO-based reinforcement learning to refine intrinsic reasoning, and (3) intuitive calibration that translates intrinsic reasoning into stable and calibrated final decisions.}
    \label{fig:pipeline}
    \vspace{-15pt}
\end{figure}

\section{Method}
\label{sec:method}

\begin{figure}[!t]
    \centering
    \includegraphics[width=\textwidth]{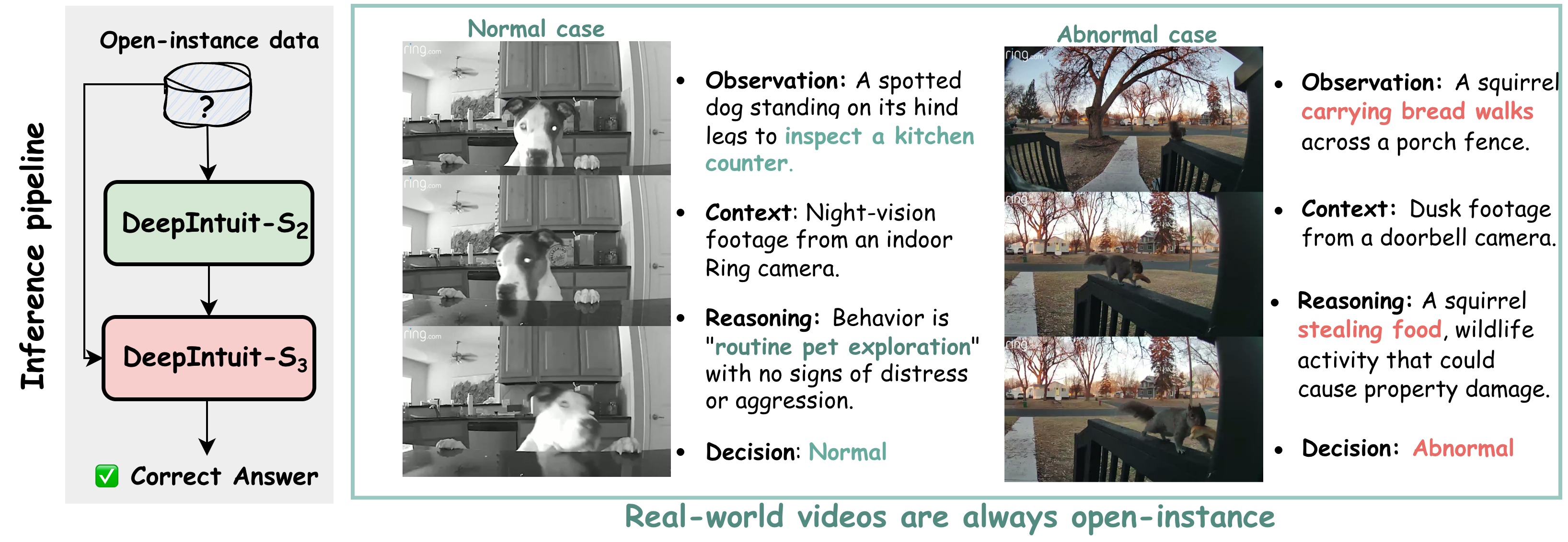}
    \caption{\textbf{Qualitative examples on open-instance videos.} The refined model generates structured intrinsic reasoning (\textit{e.g.}, observations and context) before predicting the final label. The examples show accurate classification of both normal and abnormal events, illustrating robust open-instance generalization in real-world scenarios.}
    \label{fig:infer}
    \vspace{-5pt}
\end{figure}

{
We begin with the problem formulation in Sec.~\ref{sec:problem}, and then present an intrinsic reasoning framework that evolves open-instance video classification from imitation to intuition. Rather than using a vision-language model as a direct classifier, the framework develops its latent reasoning ability and converts it into reliable classification behavior through three stages: \textbf{1)} cold-start supervised alignment (Sec.~\ref{sec:cold}), \textbf{2)} GRPO-based reinforcement learning (Sec.~\ref{sec:rl}), and \textbf{3)} intuitive calibration (Sec.~\ref{sec:caliberation}). Together, these stages establish reasoning priors, refine the reasoning process, and produce stable and calibrated decisions. The overall training pipeline is illustrated in Figure~\ref{fig:pipeline}, while the inference pipeline and representative examples are shown in Figure~\ref{fig:infer}.
}

\subsection{Formulation}
\label{sec:problem}

{
We consider an open-instance video classification task, where each input video $x \in \mathcal{X}$ is associated with a label $y \in \mathcal{Y}$. A traditional classifier directly predicts the label from the input, $\hat{y} = f(x)$, but such a direct input-to-label mapping is often brittle in open-instance settings, where classification requires richer semantic understanding under large intra-class variation.
}

{
To overcome this, we formulate classification through intrinsic reasoning. A VLM first produces an intrinsic reasoning trace $R$ and a provisional prediction $\hat{y}_r$:
\begin{equation}
    (R, \hat{y}_r) = g(x),
\end{equation}
where $g$ is initialized by cold-start supervised alignment and further enhanced by GRPO-based reinforcement learning. We then introduce an intuitive calibration module $h$, which maps the input, the generated reasoning trace, and the provisional prediction to the final output:
\begin{equation}
    \hat{y} = h(x, R, \hat{y}_r).
\end{equation}
The calibration module is trained on reasoning traces generated by the same refined model, so that reasoning and decision making remain distribution-consistent. The overall prediction process is:
\begin{equation}
    \hat{y} = h\big(x, g(x)\big),
\end{equation}
where intrinsic reasoning is treated as an intermediate representation, and final classification is obtained through explicit calibration.
}

\subsection{Reasoning Initialization and Enhancement}
\label{sec:transfer}

{
The first two stages of \model focus on intrinsic reasoning initialization and enhancement, where the model learns to produce structured reasoning traces together with provisional predictions for open-instance video classification. Given an input video $x \in \mathcal{X}$, the reasoning model $g_{\theta}$ defines a conditional distribution
\[
(R, \hat{y}_r) \sim g_{\theta}(R, \hat{y}_r \mid x),
\]
where $R=(r_1,\dots,r_T)$ denotes an intrinsic reasoning trace and $\hat{y}_r \in \mathcal{Y}$ is the corresponding provisional prediction.
}

\begin{figure}[!t]
    \centering
    \includegraphics[width=\textwidth]{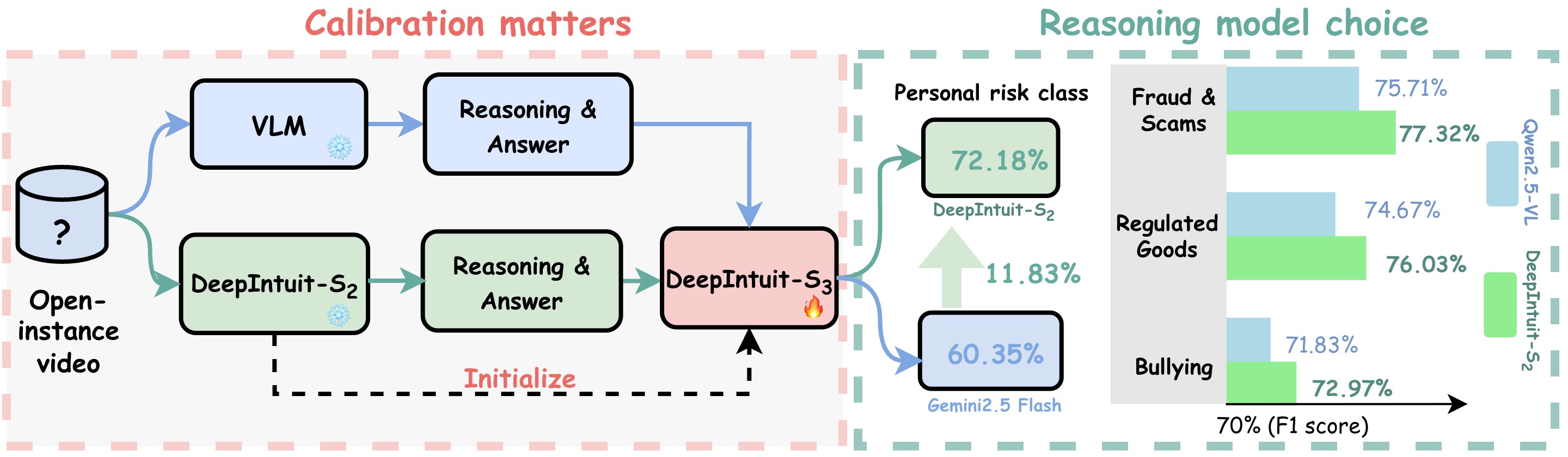}
    \caption{\textbf{Effect of calibration and reasoning model choice.} \textbf{Left:} Initializing Stage-3 from the Stage-2 model yields a $>10\%$ F1 improvement compared with using an external VLM model. \textbf{Right:} DeepIntuit-S$_2$, trained with cold-start supervised alignment and GRPO, consistently outperforms the baseline reasoning model (\textit{e.g.}, Qwen2.5-VL) across categories.}
    \label{fig:self_reasoning}
    \vspace{-10pt}
\end{figure}

{
\subsubsection{Cold-start Supervised Alignment.}
\label{sec:cold}
Direct reinforcement learning over long reasoning trajectories is often unstable due to sparse rewards and the large output space. We therefore first initialize $g_{\theta}$ with a cold-start dataset
\[
\mathcal{D}_{\text{cs}}=\{(x_i, R_i^{*}, \hat{y}_{r,i}^{*})\}_{i=1}^{N},
\]
where the reasoning traces are generated by a teacher model with reasoning ability. The model is first optimized with supervised learning:
\[
\mathcal{L}_{\text{cs}}(\theta)
= - \mathbb{E}_{(x,R^{*},\hat{y}_r^{*}) \sim \mathcal{D}_{\text{cs}}}
\big[\log g_{\theta}(R^{*}, \hat{y}_r^{*} \mid x)\big],
\]
which establishes an initial reasoning prior and provides a stable starting point for subsequent reinforcement learning.
}

{
\subsubsection{GRPO-based Refinement.}
\label{sec:rl}
After cold-start alignment for initialization, we further refine the reasoning model using Group Relative Policy Optimization (GRPO)~\cite{grpo_cite}. For each input $x$, we sample a group of $K$ candidate reasoning trajectories
\[
\{(R^{(k)}, \hat{y}_r^{(k)})\}_{k=1}^{K} \sim g_{\theta}(\cdot \mid x),
\]
and assign each a scalar reward $\mathcal{R}^{(k)}=\mathcal{R}(x,R^{(k)},\hat{y}_r^{(k)})$, computed by rule-based evaluators that measure reasoning quality and prediction correctness. The optimization objective is
\[
\mathcal{L}_{\text{GRPO}}(\theta)
= - \mathbb{E}_{x}
\left[
\sum_{k=1}^{K}
\tilde{w}^{(k)} \log g_{\theta}(R^{(k)}, \hat{y}_r^{(k)} \mid x)
\right],
\]
where the normalized weights are
\[
\tilde{w}^{(k)}
=
\frac{\exp(\mathcal{R}^{(k)}/\tau)}
{\sum_{j=1}^{K}\exp(\mathcal{R}^{(j)}/\tau)},
\]
and $\tau$ is a temperature hyperparameter.
This stage further improves the reasoning process by encouraging more coherent and discriminative reasoning traces. We name above two processes as intrinsic reasoning. The resulting model produces stronger provisional predictions, while the final classification is still deferred to the intuitive calibration stage.
}

\subsection{Intrinsic Reasoning with Calibration}
\label{sec:caliberation}

{
While the enhanced reasoning model produces informative intrinsic reasoning traces, its provisional predictions are not always reliable enough for final classification. To obtain stable and calibrated decisions, we introduce an intuitive calibration stage that explicitly decouples decision making from reasoning generation.
}

{
Given an input video $x$, the trained reasoning model $g_{\theta}$ produces an intrinsic reasoning trace $R$ and a provisional prediction $\hat{y}_r$. The calibration module $h_{\phi}$ then predicts the final label by conditioning on both the original input and the generated reasoning:
\[
\hat{y} = h_{\phi}(x, R, \hat{y}_r),
\]
where $h_{\phi}$ outputs a calibrated prediction over the label space $\mathcal{Y}$.
}

{
Then, \model is trained by supervised learning using intrinsic reasoning traces and we name this process as calibration.
Specifically, \model is trained on tuples
\[
\mathcal{D}_{\text{cal}} = \{(x_i, R_i, \hat{y}_{r,i}, y_i)\}_{i=1}^{M},
\]
where $(R_i, \hat{y}_{r,i})$ are generated by the frozen refined reasoning model $g_{\theta}$. The training objective is the standard cross-entropy loss:
\[
\mathcal{L}_{\text{cal}}(\phi)
= - \mathbb{E}_{(x,R,\hat{y}_r,y) \sim \mathcal{D}_{\text{cal}}}
\big[ \log h_{\phi}(y \mid x, R, \hat{y}_r) \big].
\]
By training directly on intrinsic reasoning traces generated by the same enhanced model, the calibration module preserves distribution consistency between reasoning and decision making. It learns when to rely on the generated reasoning and when to correct it, rather than treating reasoning outputs as final evidence. Since $h_{\phi}$ is optimized with supervised objectives, it inherits the stability and calibration properties of standard classifiers while benefiting from intrinsic reasoning as an intermediate representation. This design avoids common failure modes of directly using reasoning outputs for classification, including overconfident predictions and plausible but incorrect final decisions.
}

\section{Experiment}
\label{sec:blind}

\subsection{Datasets}

{
We evaluate \model on two public benchmarks and one in-house dataset, all of which reflect challenging open-instance video classification scenarios with substantial ambiguity, semantic complexity, and large intra-class variation.
}

{
\subsubsection{SmartHome-LLM Benchmark~\cite{zhao2025smarthome}.}
SmartHome-LLM focuses on household monitoring and anomaly recognition. It contains 1,011 available real-world smart-home video clips spanning seven daily-life themes, such as wildlife intrusion and elderly assistance. Each clip is paired with structured annotations, including event labels, explanatory notes, and reasoning traces. This benchmark is particularly challenging because many abnormal events are subtle and context-dependent, while the large diversity of home environments requires models to go beyond simply visual feature matching and perform more robust semantic understanding. 
} We divide the dataset into 815 training samples and 196 test samples. All methods are evaluated under this setting.
\newline\noindent{
\textbf{MultiHateClip~\cite{wang2024multihateclip}.}
MultiHateClip is a multilingual benchmark for harmful video content detection. It contains 2,000 annotated videos categorized into hateful, offensive, and benign classes. In our experiments, we use the English subset. The task is challenging because harmful intent is often expressed through the interaction of visual content, spoken language, and on-screen text, requiring multimodal semantic reasoning. In addition, the boundary between hateful and offensive content is often subtle, making reliable classification difficult under nuanced semantic variation.
}We adopt the same dataset split as~\cite{wang2024multihateclip}.
\newline\noindent{
\textbf{In-house dataset.}
We further construct a proprietary large-scale dataset for video content moderation via open-instance video classification. The dataset covers multiple safety-related categories commonly encountered on online platforms, including Frauds and Scams, Regulated Goods, Bullying, Personal Risk, among others, capturing a broad spectrum of harmful or policy-violating behaviors in real-world video content. These categories represent diverse moderation scenarios ranging from personal safety threats to policy-violating commercial activities and abusive behaviors. The dataset is built through a structured generation-and-filtering pipeline. We first use Gemini 2.5-Flash~\cite{comanici2025gemini} to generate multi-step reasoning traces together with provisional predictions, and then apply in-house labeling and filtering to improve consistency and reliability. The final dataset contains 80--130K training samples and 4.5K evaluation samples. From the training set, we derive three disjoint subsets: 10K samples for cold-start supervised alignment, 30--50K samples for GRPO-based reinforcement learning, and 40--70K samples for intuitive calibration. For the calibration stage, we further expand the data by applying multi-rollout generation (rollout number $=4$) with the GRPO-refined model, and resampling the resulting trajectories to construct 160--280K intrinsic reasoning instances. This staged construction aligns with our three-stage framework, providing stable initialization, effective reasoning refinement, and distribution-consistent calibration.
}
\subsection{Experimental Details}
\noindent\textbf{Implementation.}
Our framework is built upon Qwen2.5-VL-7B~\cite{bai2025qwen2} and employs a progressive training strategy to enhance reasoning capability. We first perform supervised fine-tuning of the vision-language backbone to establish stable multimodal alignment and basic structured reasoning behaviors, following the default configuration of Qwen2.5-VL~\cite{qwen2p5_cite}. 
We then further improve reasoning ability using Group Relative Policy Optimization (GRPO). Training follows a DAPO-style rollout and update protocol, where the policy is optimized for one epoch with a batch size of 64 and a learning rate of $2\times10^{-5}$. For each training instance, eight rollouts are sampled to estimate relative advantages, and the policy is updated once per sampling round.
Finally, the refined model is used to generate four reasoning samples per training instance to construct a structured meta-review dataset, which is then used for additional supervised fine-tuning for ten epochs under the same hyperparameter configuration.
\newline\noindent\textbf{Baselines.}
To comprehensively evaluate our approach, we compare against a diverse set of strong baselines spanning both specialized video encoders and large multimodal models. First, we include state-of-the-art video understanding architectures such as UniFormerV2~\cite{li2022uniformerv2} and InternVideo2-6B~\cite{wang2024internvideo2}, which are specifically optimized for spatiotemporal modeling and serve as competitive task-oriented baselines. We further benchmark against powerful proprietary vision-language models, including GPT-4–series~\cite{achiam2023gpt,openai2025gpt4omini,openaio1_cite} and Gemini-2.5 variants~\cite{comanici2025gemini}, which represent general-purpose multimodal reasoning systems with strong zero-shot capabilities. In addition, we report results from the Qwen-2.5-7B backbone under different training paradigms, including zero-shot inference, Direct-SFT,RB-FT~\cite{xu2025rb} and reinforcement-based fine-tuning variants~\cite{grpo_cite}. This broad comparison ensures that improvements are not limited to a particular architecture family but hold across both specialized video models and large-scale multimodal LLMs.
\newline\noindent\textbf{Evaluation Metrics.}
We report overall accuracy and class-wise F1 scores following previous methods~\cite{ho2025dejavid,xu2025rb}. Due to class imbalance in datasets such as MultiHateClip and our in-house dataset, we emphasize F1 score, which better reflect performance on minority classes and balanced recognition across categories.
\subsection{Comparison}
\begin{table*}[!t]
\centering
\caption{\textbf{Quantitative results on the MultiHateClip benchmark.} We compare \model with close-sourced VLMs, conventional video encoders, and open-sourced Qwen2.5-VL-7B variants under different post-training strategies. Here, Nor., Hat., Off., and Avg. denote Normal, Hateful, Offensive, and Average, respectively.}
\scriptsize
\label{tab:multihateclip}
\setlength{\tabcolsep}{4.5pt}
\begin{tabular}{llcccccc}
\toprule
\multirow{2}{*}{Model} & \multirow{2}{*}{Method} & \multirow{2}{*}{Open} & \multirow{2}{*}{Acc. ($\uparrow$)} & \multicolumn{4}{c}{F1 Score ($\%,\uparrow$)} \\
\cmidrule{5-8}
& & & & Nor. & Hat. & Off. & Avg. \\
\midrule
{\color{gray}\itshape GPT} 
& {\color{gray}\itshape GPT-4o~\cite{gpt4o}} 
& {\color{gray}\itshape $\checkmark$} 
& {\color{gray}\itshape 72.19} 
& {\color{gray}\itshape 84.16} 
& {\color{gray}\itshape 55.45} 
& {\color{gray}\itshape 12.50} 
& {\color{gray}\itshape 50.70} \\
{\color{gray}\itshape GPT} 
& {\color{gray}\itshape GPT-4.1~\cite{achiam2023gpt}} 
& {\color{gray}\itshape $\checkmark$} 
& {\color{gray}\itshape 70.46} 
& {\color{gray}\itshape 83.58} 
& {\color{gray}\itshape 60.38} 
& {\color{gray}\itshape 23.53} 
& {\color{gray}\itshape 55.83} \\
\cmidrule{2-8}
{\color{gray}\itshape Gemini} 
& {\color{gray}\itshape Gemini-2.5-Flash~\cite{comanici2025gemini}} 
& {\color{gray}\itshape $\checkmark$} 
& {\color{gray}\itshape 70.94} 
& {\color{gray}\itshape 83.98} 
& {\color{gray}\itshape 18.18} 
& {\color{gray}\itshape 40.00} 
& {\color{gray}\itshape 47.39} \\
{\color{gray}\itshape Gemini} 
& {\color{gray}\itshape Gemini-2.5-Pro~\cite{comanici2025gemini}} 
& {\color{gray}\itshape $\checkmark$} 
& {\color{gray}\itshape 75.81} 
& {\color{gray}\itshape 84.71} 
& {\color{gray}\itshape 61.11} 
& {\color{gray}\itshape 0.00} 
& {\color{gray}\itshape 48.61} \\
\midrule
\multirow{2}{*}{Video Encoder} 
& UniFormerV2~\cite{li2022uniformerv2} & $\times$ & 59.17 & 72.10 & 8.33  & 35.42 & 38.62 \\
& InternVideo2-6B~\cite{wang2024internvideo2} & $\times$ & 62.72 & 74.85 & 10.91 & 38.06 & 41.27 \\
\midrule
\multirow{5}{*}{Qwen2.5-VL-7B~\cite{bai2025qwen2} } 
& Zero-shot & $\checkmark$ & 54.48 & 66.90 & 2.70  & 40.12 & 36.57 \\
& Direct-SFT & $\checkmark$ & 66.86 & 79.92 & 11.11 & 41.35 & 44.13 \\
& RB-FT~\cite{xu2025rb} & $\checkmark$ & 71.00 & 83.47 & 23.53 & 49.78 & 52.26 \\
& SFT+GRPO~\cite{grpo_cite} & $\checkmark$ & 67.88 & 81.78 & 0.00  & 33.85 & 38.54 \\
& \textbf{\model} & \textbf{$\checkmark$} & \textbf{72.72} & \textbf{83.49} & \textbf{30.00} & \textbf{56.52} & \textbf{56.67} \\
\bottomrule
\end{tabular}
\end{table*}

{
We organize the comparisons on both MultiHateClip and SmartHome into three groups: (2) close-sourced large multimodal models, \textit{i.e.}, GPT and Gemini series; (2) traditional video encoder models trained only on the target training set without external reasoning priors;  and (3) open-sourced Qwen2.5-VL~\cite{bai2025qwen2} variants under different post-training strategies.
}

{
\subsubsection{Comparison with close-sourced VLMs.}
Proprietary models, including GPT-4~\cite{achiam2023gpt} variants and Gemini-2.5~\cite{comanici2025gemini} (Flash/Pro), demonstrate strong zero-shot generalization, benefiting from large-scale multimodal pretraining and stronger semantic priors. For example, Gemini-2.5-Pro achieves competitive performance on several metrics. However, despite their strong overall capability, these black-box systems still exhibit inconsistent gains on the most challenging categories, and their reasoning behavior cannot be explicitly refined or calibrated for our target setting.
}

{
\subsubsection{Comparison with video encoder models.}
Representative video backbones such as InternVideo2-6B~\cite{wang2024internvideo2} and UniFormerV2~\cite{li2022uniformerv2} rely primarily on supervised learning within the target training distribution. While they perform competitively in relatively homogeneous settings, their performance degrades in open-instance scenarios with substantial intra-class variation. As shown in Table~\ref{tab:multihateclip} and Table~\ref{tab:smarthome}, on both MultiHateClip and SmartHome, these models show limited robustness and clear trade-offs across class-wise F1 scores, especially in semantically ambiguous or safety-sensitive cases.
}

\begin{table*}[!t]
\centering
\caption{\textbf{Quantitative results on the SmartHome-LLM benchmark.} \model achieves the best results on all reported metrics, including overall accuracy, class-wise F1, and average F1, demonstrating stronger and more balanced performance on both Normal and Abnormal events.}
\scriptsize
\label{tab:smarthome}
\setlength{\tabcolsep}{5.3pt}
\begin{tabular}{llccccc}
\toprule
\multirow{2}{*}{Model} & \multirow{2}{*}{Method} & \multirow{2}{*}{Open} & \multirow{2}{*}{Acc. ($\uparrow$)} & \multicolumn{3}{c}{F1 Score ($\%,\uparrow$)} \\
\cmidrule{5-7}
& & & & Normal & Abnormal & Avg. \\
\midrule
{\color{gray}\itshape GPT}
& {\color{gray}\itshape GPT-4o~\cite{gpt4o}}
& {\color{gray}\itshape $\checkmark$}
& {\color{gray}\itshape 62.76}
& {\color{gray}\itshape 63.32}
& {\color{gray}\itshape 62.18}
& {\color{gray}\itshape 62.75} \\
{\color{gray}\itshape GPT}
& {\color{gray}\itshape GPT-4.1~\cite{achiam2023gpt}}
& {\color{gray}\itshape $\checkmark$}
& {\color{gray}\itshape 70.53}
& {\color{gray}\itshape 69.23}
& {\color{gray}\itshape 71.72}
& {\color{gray}\itshape 70.48} \\
\cmidrule{2-7}
{\color{gray}\itshape Gemini}
& {\color{gray}\itshape Gemini-2.5-Flash~\cite{comanici2025gemini}}
& {\color{gray}\itshape $\checkmark$}
& {\color{gray}\itshape 73.33}
& {\color{gray}\itshape 70.79}
& {\color{gray}\itshape 75.47}
& {\color{gray}\itshape 73.13} \\
{\color{gray}\itshape Gemini}
& {\color{gray}\itshape Gemini-2.5-Pro~\cite{comanici2025gemini}}
& {\color{gray}\itshape $\checkmark$}
& {\color{gray}\itshape 73.47}
& {\color{gray}\itshape 70.11}
& {\color{gray}\itshape 76.15}
& {\color{gray}\itshape 73.13} \\
\midrule
\multirow{2}{*}{Video Encoder}
& UniFormerV2~\cite{li2022uniformerv2} & $\times$ & 70.12 & 38.02 & 78.95 & 58.49 \\
& InternVideo2-6B~\cite{wang2024internvideo2} & $\times$ & 68.70 & 35.40 & 79.10 & 57.25 \\
\midrule
\multirow{4}{*}{Qwen2.5-VL-7B~\cite{bai2025qwen2}}
& Zero-shot & $\checkmark$ & 55.10 & 60.36 & 48.24 & 54.30 \\
& Direct-SFT & $\checkmark$ & 76.02 & 51.55 & 84.07 & 67.81 \\
& RB-FT~\cite{xu2025rb} & $\checkmark$ & 82.65 & 76.06 & 86.40 & 81.23 \\
& \textbf{\model} & \textbf{$\checkmark$} & \textbf{88.27} & \textbf{83.45} & \textbf{90.91} & \textbf{87.18} \\
\bottomrule
\end{tabular}
\end{table*}

{
\subsubsection{Comparison with post-training methods.}
We further compare against Qwen2.5-VL-7B~\cite{bai2025qwen2} under several post-training strategies, including zero-shot inference, direct supervised fine-tuning (Direct-SFT), reinforcement-based fine-tuning (RL-FT~\cite{grpo_cite}), and the two-stage CoT fine-tuning strategy RB-FT~\cite{xu2025rb}. While these methods improve over the zero-shot baseline, their gains remain limited and often uneven across classes. On SmartHome, post-training improves overall accuracy but still struggles to maintain strong performance on both Normal and Abnormal classes simultaneously. On MultiHateClip, category-wise improvements remain modest, particularly for difficult classes such as Offensive.
}

{
Built on Qwen2.5-VL-7B, our \model outperforms most of these three groups across both benchmarks. On MultiHateClip, it achieves 72.72\% overall accuracy and the highest F1 score in the Offensive category (56.52\%), demonstrating stronger robustness under semantic ambiguity. On SmartHome, \model establishes new state-of-the-art results in both overall accuracy and average F1, while achieving more balanced performance between Normal and Abnormal classes.
}

\subsection{Ablation Study \& Analysis}
\noindent\textbf{Intrinsic Reasoning Improves Robustness.}
Figure~\ref{fig:self_reasoning} shows the effect of intrinsic reasoning in open-instance video classification. Instead of directly distilling final labels from a VLM teacher, we initialize the student with reasoning traces and train it to make predictions through an explicit reasoning process. This reasoning-driven transfer consistently improves performance across challenging categories, yielding F1 gains of 1.61\% on Scams, 1.36\% on Regulated Goods, and 1.14\% on Bullying. These results suggest that intrinsic reasoning provides a stronger transfer signal than direct answer imitation, leading to better robustness under large intra-class variation and semantically ambiguous cases.

\begin{table*}[!t]
    \centering
    \begin{minipage}{0.52\textwidth}
        \centering
        \resizebox{\textwidth}{!}{
        \begin{tabular}{llcc}
        \toprule
        Category & Setting & Imitation & Calibration \\
        \midrule
        \multicolumn{4}{l}{\textit{(a) Initialization strategy}} \\
        \midrule
        \multirow{2}{*}{Group\#1}
        & CoT  & 55.60 & 67.10 \\
        & GRPO & 60.49 & \textbf{67.22} \\
        \midrule
        \multirow{2}{*}{Group\#2}
        & CoT  & 65.37 & 67.86 \\
        & GRPO & 62.38 & \textbf{68.81} \\
        \midrule
        \multicolumn{4}{l}{\textit{(b) Backbone choice}} \\
        \midrule
        \multirow{3}{*}{Group\#1}
        & VLM$_{\text{in-house v1}}$ & 60.49 & 67.22 \\
        & VLM$_{\text{in-house v2}}$ & 61.97 & 70.40 \\
        & VLM$_{\text{in-house v3}}$ & 70.40 & \textbf{75.71} \\
        \midrule
        \multirow{3}{*}{Group\#2}
        & VLM$_{\text{in-house v1}}$ & 62.38 & 68.81 \\
        & VLM$_{\text{in-house v2}}$ & 70.11 & 72.35 \\
        & VLM$_{\text{in-house v3}}$ & 72.35 & \textbf{74.67} \\
        \bottomrule
        \end{tabular}}
        \vspace{12pt}
        \caption{\textbf{Ablation on initialization and backbones.} F1 (\%) comparison of CoT vs.\ GRPO initialization and different backbones, reported before and after calibration.}
        \label{tab:imitation}
    \end{minipage}
    \begin{minipage}{0.44\textwidth}
        \centering
        \includegraphics[width=\textwidth]{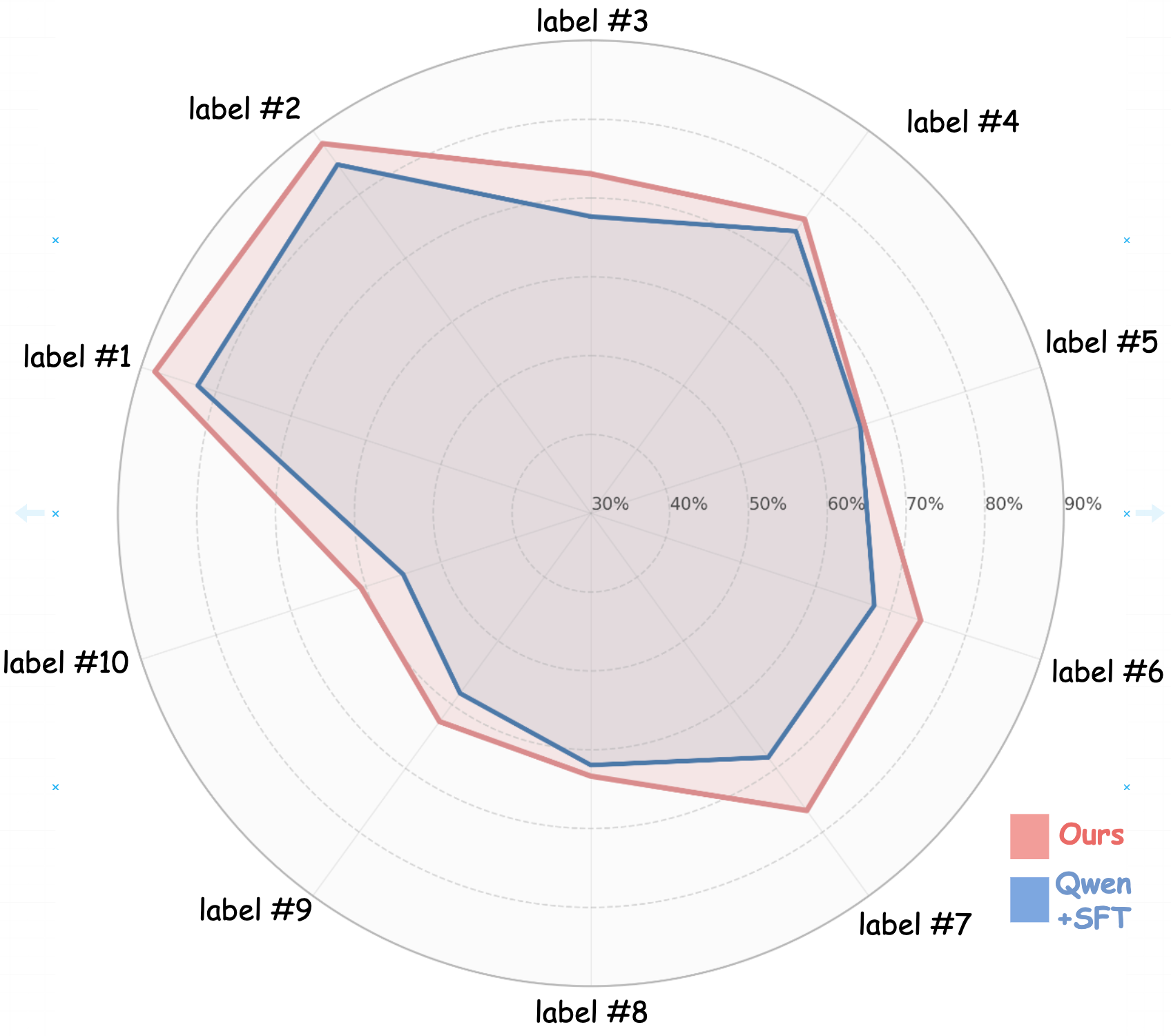}
        \vspace{5pt}
        \captionof{figure}{\textbf{Category-wise impact.} Class-wise performance of \model versus Qwen+SFT, showing consistent gains across categories.}
        \label{fig:class}
    \end{minipage}
\end{table*}

\subsubsection{Largest Gains on Semantically Challenging Categories.}
We further analyze category-wise performance to understand where the proposed framework is most beneficial. As shown in Figure~\ref{fig:class}, our method consistently improves over the baseline across most categories. The largest gains appear on label \#6 and label \#7, where our approach significantly outperforms the Qwen-SFT baseline, indicating that intrinsic reasoning is particularly beneficial for categories requiring more contextual interpretation. Noticeable improvements are also observed on label \#1 and label \#2, where the performance gap remains consistently positive. Moderate gains are achieved on label \#3, label \#8, and label \#9, suggesting improved robustness in more ambiguous scenarios. By contrast, the improvement is relatively smaller on label \#4 and label \#5, where the baseline already performs competitively and the decision boundaries may rely more on explicit visual evidence. Overall, the results demonstrate that intrinsic reasoning provides consistent benefits across diverse categories, with the most substantial improvements appearing in cases that require deeper semantic understanding.

\vspace{8pt}
\noindent\textbf{GRPO Refines Reasoning Beyond Imitation.}
\noindent\textbf{GRPO Refines Reasoning Beyond Imitation.}
The ablation in Table~\ref{tab:imitation} verifies the value of GRPO-based reinforcement learning beyond direct supervised imitation of teacher-generated reasoning traces. Compared with training only on Gemini-generated CoT traces, GRPO produces a stronger reasoning model, improving accuracy on Group\#1 by 4.89\%.

\begin{wrapfigure}{r}{0.45\textwidth}
\centering
\vspace{-15pt}
\includegraphics[width=\linewidth]{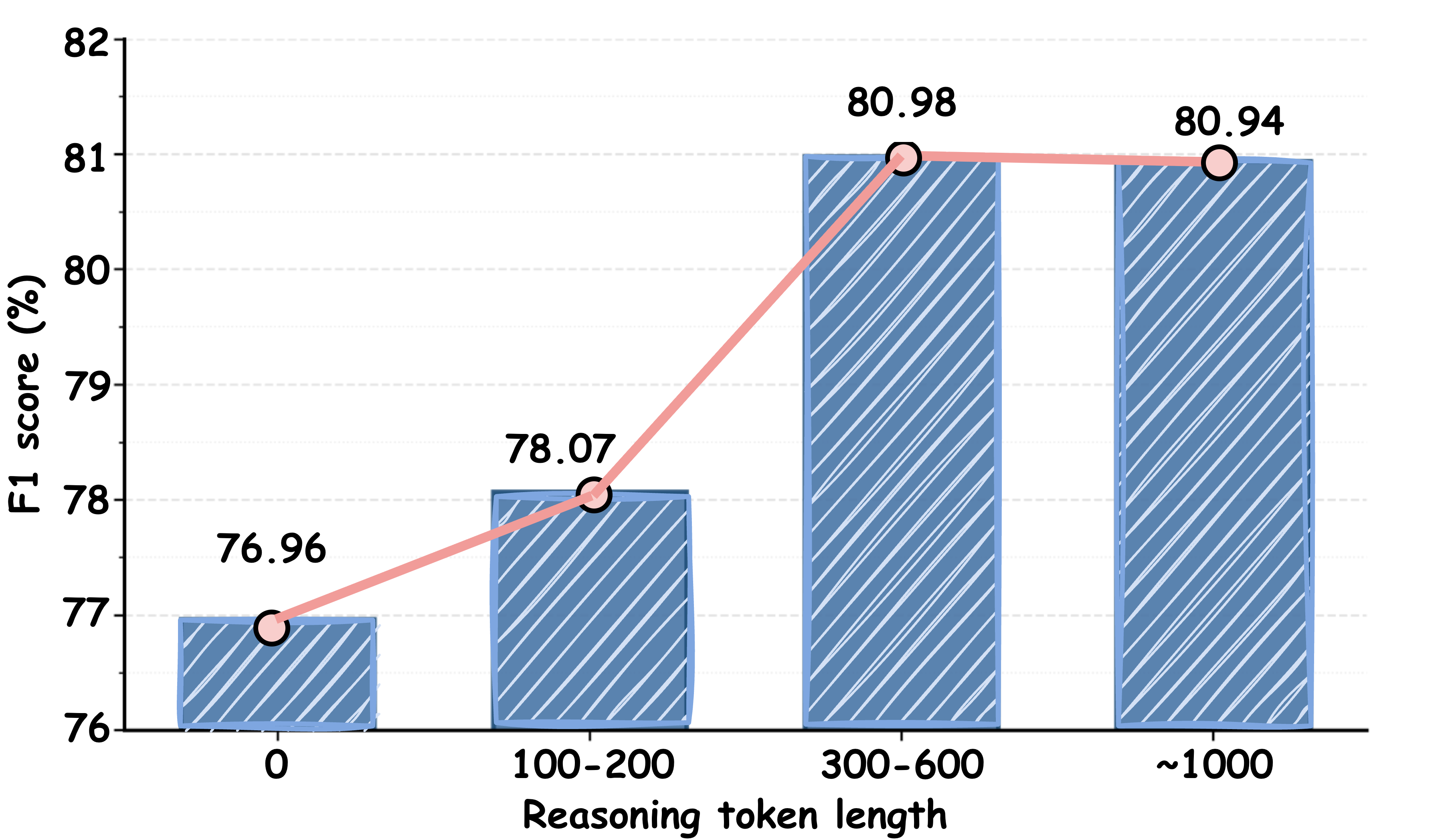}
\caption{\textbf{Effect of reasoning token length on performance.} Increasing intrinsic reasoning improves performance from short to moderate lengths, while very long reasoning yields little additional benefit.}
\label{fig:reasoning_length}
\vspace{-35pt}
\end{wrapfigure}
More importantly, this improvement is not limited to intermediate reasoning quality: it translates into consistently better downstream performance after intuitive calibration. This suggests that GRPO does not merely mimic teacher reasoning, but further refines intrinsic reasoning in a way that provides a stronger foundation for final classification.

\subsubsection{Moderate-Length Reasoning Works Best.}
We analyze the effect of reasoning token length by progressively increasing the maximum number of generated reasoning tokens. 
As shown in Figure~\ref{fig:reasoning_length}, introducing 
explicit intrinsic reasoning consistently improves performance over the no-reasoning baseline (76.96\%). Short reasoning (100--200 tokens) yields a modest gain to 78.07\%, while medium-length reasoning (300--600 tokens) provides the largest improvement, reaching 80.98\%. Further extending the reasoning to around 1000 tokens does not bring additional gains (80.94\%), suggesting diminishing returns beyond a moderate length. These results indicate that intrinsic reasoning is most effective when it is sufficiently informative but not overly long, as excessively long reasoning may introduce redundancy without improving final classification.

\subsubsection{Stronger Backbones Unlock Larger Gains.}
{
The results in Table~\ref{tab:imitation} also show that backbone choice is critical for both reasoning development and final classification performance. In our experiments, we evaluate three internally developed vision-language models, denoted as VLM$_{\text{in-house v1}}$, VLM$_{\text{in-house v2}}$, and VLM$_{\text{in-house v3}}$, which represent progressively stronger versions of our in-house architecture with improved multimodal representation capability. Across these variants, stronger models consistently achieve higher accuracy both before and after GRPO-based reinforcement learning, indicating that better vision-language foundations provide stronger initial reasoning priors and benefit more from subsequent reasoning refinement. For example, VLM$_{\text{in-house v3}}$ substantially outperforms VLM$_{\text{in-house v1}}$ at initialization (\textit{e.g.}, 70.40\% vs.\ 60.49\% on Frauds and Scams, and 72.35\% vs.\ 62.38\% on Regulated Goods), and this advantage remains after GRPO refinement (75.71\% vs.\ 67.22\%, and 74.67\% vs.\ 68.81\%, respectively). Moreover, the improvement obtained from reasoning refinement is larger for stronger models, suggesting that models with richer semantic priors can more effectively exploit the proposed training framework. This trend indicates a positive interaction between backbone capability and reasoning optimization. Overall, these results demonstrate that while our method is compatible with different vision-language backbones, its full potential is best realized when applied to stronger multimodal foundation models.
}

\section{Conclusion}

{
In this paper, we study how reinforcement learning--enhanced reasoning can be effectively used for open-instance video classification. We show that directly deploying RL-refined reasoning models remains brittle, because stronger reasoning does not automatically produce reliable or calibrated final decisions. To address this, we propose \model, an intrinsic reasoning framework that evolves open-instance video classification from imitation to intuition through three stages: cold-start supervised alignment, GRPO-based reinforcement learning, and intuitive calibration. By explicitly decoupling reasoning generation from final decision making, \model develops latent reasoning ability and translates it into stable classification behavior without directly treating reasoning outputs as final evidence. Experiments on diverse and challenging benchmarks show that this design leads to stronger robustness and generalization, and that distribution-consistent calibration is critical for stable performance under large intra-class variation.
}

\bibliographystyle{splncs04}
\bibliography{egbib}
\end{document}